\documentclass{article}

\PassOptionsToPackage{numbers, compress}{natbib}

\usepackage[preprint]{neurips_2023}

\usepackage[utf8]{inputenc} 
\usepackage[T1]{fontenc}    
\usepackage{hyperref}       
\usepackage{url}            
\usepackage{booktabs}       
\usepackage{amsfonts}       
\usepackage{nicefrac}       
\usepackage{microtype}      
\usepackage{xcolor}         
\usepackage{tcolorbox}
\usepackage{array}
\newcolumntype{L}{>{\centering\arraybackslash}m{3cm}}
\definecolor{blue}{HTML}{486aa1} 
\definecolor{red}{HTML}{de5454} 
\usepackage{longtable}
\usepackage[shortlabels]{enumitem}

\title{(Ir)rationality and Cognitive Biases in Large Language Models}

\author{%
  Olivia Macmillan-Scott\thanks{Corresponding author.}\\
  University College London\\
  \texttt{olivia.macmillan-scott.16@ucl.ac.uk} \\
  \And
  Mirco Musolesi \\
  University College London\\
  University of Bologna \\
  \texttt{m.musolesi@ucl.ac.uk} \\
}

\begin{document}

\maketitle

\begin{abstract}
Do large language models (LLMs) display rational reasoning? LLMs have been shown to contain human biases due to the data they have been trained on; whether this is reflected in rational reasoning remains less clear. In this paper, we answer this question by evaluating seven language models using tasks from the cognitive psychology literature. We find that, like humans, LLMs display irrationality in these tasks. However, the way this irrationality is displayed does not reflect that shown by humans. When incorrect answers are given by LLMs to these tasks, they are often incorrect in ways that differ from human-like biases. On top of this, the LLMs reveal an additional layer of irrationality in the significant inconsistency of the responses. Aside from the experimental results, this paper seeks to make a methodological contribution by showing how we can assess and compare different capabilities of these types of models, in this case with respect to rational reasoning.
\end{abstract}

\section{Introduction}

Large language models (LLMs) have quickly become integrated into everyday activities, and their increasing capabilities mean this will only become more pervasive. Given this notion, it is important for us to develop methodologies to evaluate the behaviour of LLMs. As we will see, these models still exhibit biases and produce information that is not factual \cite{chang_2023}. However, there is extensive variation in the responses given by different models to the same prompts. In this paper, we take a comparative approach based in cognitive psychology to evaluate the rationality and cognitive biases present in a series of LLMs; the aim of this paper is to provide a method to evaluate and compare the behaviour and capabilities of different models, here with a focus on rational and irrational reasoning. There exist different definitions of what is rational in artificial intelligence \cite{russell_2016}, and conceptions vary depending on whether we are looking at reasoning or behaviour \cite{macmillanscott_2023}. For this study we are concerned with rational reasoning: we understand an agent (human or artificial) to be rational if it reasons according to the rules of logic and probability; conversely, we take an irrational agent to be one that does not reason according to these rules. This is in line with Stein's \cite{stein_1996} formal definition of the \textit{Standard Picture} of rationality.

In this paper, we evaluate seven LLMs using cognitive tests proposed by Kahneman and Tversky \cite{kahneman_tversky_1972,tversky_kahneman_1974,tversky_kahneman_1983} and others \cite{wason_1966,eddy_1982,friedman_1998}, as well as some facilitated versions formulated by Bruckmaier et al. \cite{bruckmaier_2021}, and evaluate the responses across two dimensions: \textit{correct} and \textit{human-like} \cite{binz_schulz_2023}. These tasks were initially designed to illustrate cognitive biases and heuristics in human reasoning, showing that humans often do not reason rationally \cite{kahneman_tversky_1982}; in this case, we use them to evaluate the \textit{rationality} of LLMs. The 'holy grail' would be to develop a set of benchmarks that can be used to test the rationality of a model; this is a complex problem which requires a consensus what is deemed rational and irrational.  

In using methods designed to evaluate human reasoning, it is important to acknowledge the performance vs. competence debate \cite{firestone_2020}. This line of argument encourages \textit{species-fair} comparisons between humans and machines, meaning that we should design tests specific to either humans or machines, as otherwise apparent failures may not reflect underlying capabilities but only superficial differences. Lampinen\cite{lampinen_2023} discusses this problem when it comes to language models in particular, highlighting that different approaches must be taking to evaluate cognitive and foundation models. However, if we take the purpose of LLMs to be to produce human-like language, perhaps the best approach is precisely to evaluate their output with tasks designed to evaluate humans. This is the approach we have taken in this paper - in order to identify whether LLMs reason rationally, or whether they exhibit biases that can be assimilated to those present in human decision-making, the most appropriate approach is therefore to use tasks that were initially designed for humans. 

Building on this debate and looking at LLMs being evaluated using human tests,  Hagendorff et al. \cite{hagendorff_2023} have proposed the creation of a new field of research called \textit{machine psychology}, which would treat LLMs as participants in psychological experiments. The approach employed in this paper precisely applies tests from psychology that were originally designed for humans, in this case to evaluate rational and irrational reasoning displayed but such models. Further to this, some have even discussed the potential of using LLMs as participants in cognitive experiments instead of humans \cite{dillion_2023}, although some see this proposal as too optimistic \cite{harding_2023}, and others warn against excessive anthropomorphism \cite{rahwan_2019}. One argument against the use of such models in cognitive experiments is that LLMs may be effective at approximating average human judgements, but are not good at capturing the variation in human behaviour \cite{santurkar_2023}. One potential avenue to address this issue is current work on language models impersonating different roles \cite{salewski_2023}, in this way capturing some of the variation in human behaviour. Binz and Schulz \cite{binz_schulz_2023_b} show that after finetuning LLMs on data from psychological experiments, they can become accurate cognitive models, which they claim begins paving the way for the potential of using these models to study human behaviour. Park et al. \cite{park_2023} combine large language models with computational interactive agents to simulate human behaviour, both individual and within social settings.

Given the data that they are trained on, LLMs naturally contain human-like biases \cite{schramowski_2022,acerbi_2023,durt_2023}. Schramowski et al. \cite{schramowski_2022} highlight that language models reflect societal norms when it comes to ethics and morality, meaning that these models contain human-like biases regarding what is right and wrong. Similarly, Durt et al. \cite{durt_2023} discuss the clichés and biases exhibited by LLMs, emphasising that the presence of these biases is not due to the models' mental capacities but due to the data they are trained on. Others have focused on specific qualities of human decision-making that are not possessed by LLMs, namely the ability to reflect and learn from mistakes, and propose an approach using verbal reinforcement to address this limitation \cite{shinn_2023}. As these studies show, LLMs display human-like biases which do not arise from the models' ability to reason, but from the data they are trained on. Therefore, the question is whether LLMs also display biases that relate to reasoning: do LLMs simulate human cognitive biases? There are cases where is may be beneficial for AI systems to replicate human cognitive biases, in particular for applications that require human-AI collaboration \cite{gulati_2023}. 

To answer this question, we use tasks from the cognitive psychology literature designed to test human cognitive biases, and apply these to a series of LLMs to evaluate whether they display rational or irrational reasoning. The capabilities of these models are quickly advancing, therefore the aim of this paper is to provide a methodological contribution showing how we can assess and compare LLMs. A number of studies have taken a similar approach, however they do not generally compare across different model types \cite{binz_schulz_2023,hagendorff_2023,lamprinidis_2023,dasgupta_2023,holterman_2023,freund_2023,chen_2023,webb_2023,han_2024}, or those that do are not evaluating rational reasoning \cite{ruis_2023}. Some find that LLMs outperform humans on reasoning tasks \cite{hagendorff_2023,bubeck_2023}, others find that these models replicate human biases \cite{dasgupta_2023,itzhak_2023}, and finally some studies have shown that LLMs perform much worse than humans on certain tasks \cite{ruis_2023}. Binz and Schulz \cite{binz_schulz_2023} take a similar approach to that presented in this paper, where they treat GPT-3 as a participant in a psychological experiment to assess its decision-making, information search, deliberation and causal reasoning abilities. They assess the responses across two dimensions, looking at whether GPT-3's output is correct and/or human-like; we follow this approach in this paper as it allows us to distinguish between answers that are incorrect due to a human-like bias or are incorrect in a different way. While they find that GPT-3 performs as well or even better than human subjects, they also find that small changes to the wording of tasks can dramatically decrease the performance, likely due to GPT-3 having encountered these tasks in training. Hagendorff et al. \cite{hagendorff_2023} similarly use the Cognitive Reflection Test (CRT) and semantic illusions on a series of OpenAI's Generative Pre-trained Transformer (GPT) models. They classify the responses as \textit{correct, intuitive} (but incorrect), and \textit{atypical} - as models increase in size, the majority of responses go from being atypical, to intuitive, to overwhelmingly correct for GPT-4, which no longer displays human cognitive errors. Other studies that find the reasoning of LLMs to outperform that of humans includes Chen et al.’s \cite{chen_2023} assessment of the economic rationality of GPT, and Webb et al.’s \cite{webb_2023} comparison of GPT-3 and human performance on analogical tasks.

As mentioned, some studies have found that LLMs replicate cognitive biases present in human reasoning, and so in some instances display irrational thinking in the same way that humans do. Itzhak et al. \cite{itzhak_2023} focus on the effects of fine-tuning; they show that instruction tuning and reinforcement learning from human feedback, while improving the performance of LLMs, can also cause these models to express cognitive biases that were not present or less expressed before these fine-tuning methods were applied. While said study \cite{itzhak_2023} investigate three cognitive biases that lead to irrational reasoning, namely the decoy effect, certainty effect and belief bias, Dasgupta et al. \cite{dasgupta_2023} centre their research on the content effect and find that, like humans, models reason more effectively about believable situations than unrealistic or abstract ones. In few-shot task evaluation, the performance of LLMs is shown to increase after being provided with in-context examples, just as examples improve learning in humans \cite{lampinen_2022}. Others have found LLMs to perform worse than human subjects on certain cognitive tasks, Ruis et al. \cite{ruis_2023} test the performance of four categories of models on an \textit{implicature} task, showing that the models that perform best are those that have been fine-tuned on example-level instructions, both at the zero-shot and few-shot levels. However, they still find that models perform close to random, particularly in zero-shot evaluation. Looking at performance on mathematical problems in particular, GPT-4 has shown inconsistencies in its capabilities, correctly answering difficult mathematical questions in some instances, while also making very basic mistakes in others \cite{bubeck_2023}. As we will see below, we find this to be the case in our analysis across the language models evaluated. The inconsistency in performance is not only present in tasks involving mathematical calculations, but is apparent across the battery of tasks.

This paper forms part of the existing area of research on the evaluation of LLMs. It differs from existing work by focusing on rational and irrational reasoning, and comparing the performance of different models. As we have seen, past studies have applied cognitive psychology to study LLMs. While they often focus on seeing whether LLMs replicate different aspects of human behaviour and reasoning, such as cognitive biases, we are interested in whether the way LLMs display rational or irrational reasoning. Much of the existing work focuses on a single model, or different versions of the same model. In this case, we compare across model types and propose a way to evaluate the performance of LLMs, which may ultimately lead to the development of a set of benchmarks to test the rationality of a model.

\section{Methods}

\subsection{Language Models}

We evaluate the rational reasoning of seven LLMs using a series of tasks from the cognitive psychology literature. The models that we assess are OpenAI's GPT-3.5 \cite{brown_2020} and GPT-4 \cite{openai_2023}, Google's Bard powered by LaMDA \cite{thoppilan_2022}, Anthropic's Claude 2 \cite{anthropic_2023}, and three versions of Meta's Llama 2 model: the 7 billion (7b), 13 billion (13b) and 70 billion (70b) parameter versions \cite{touvron_2023}. We use the OpenAI API to prompt GPT-3.5 and GPT-4, and all other models are accessed through their online chatbot interfaces. The code for the former is available on GitHub, and information on how models were accessed is detailed in Appendix 1. 

We did not change any parameter settings in order to evaluate the models on these cognitive tasks. However, for Llama 2, the 7b and 13b parameter models had the default prompt shown in Figure \ref{default}. After running an initial set of the tasks on these Llama 2 models, we removed the default prompt as it generally meant that the models refused to provide a response due to ethical concerns. Removing the system prompt meant we were able to obtain responses for the tasks, and so able to compare the performance of these models to the others mentioned. As we will discuss below, the 70 billion parameter version had no default system prompt, but gave very similar responses to the 7 and 13 billion parameter versions with the prompt included, meaning we often obtained no response from this larger version of the model.

\begin{figure}[h]
\begin{tcolorbox}[width=\textwidth,colback={white},title={System prompt - Llama 2 7b and 13b},colbacktitle=red,coltitle=white]    
You are a helpful, respectful and honest assistant. Always answer as helpfully as possible, while being safe.  Your answers should not include any harmful, unethical, racist, sexist, toxic, dangerous, or illegal content. Please ensure that your responses are socially unbiased and positive in nature. If a question does not make any sense, or is not factually coherent, explain why instead of answering something not correct. If you don't know the answer to a question, please don't share false information.
\end{tcolorbox}   
\caption{Default system prompt for Llama 2 7b and 13b.} 
\label{default}
\end{figure}

\subsection{Description of Tasks}

The tasks used to evaluate these models are taken primarily from Kahneman and Tversky's work \cite{kahneman_tversky_1972,tversky_kahneman_1974,kahneman_tversky_1982,tversky_kahneman_1983}, who designed a series of tasks to highlight biases and heuristics in human reasoning. Additional tasks \cite{wason_1966,eddy_1982,friedman_1998} and facilitated versions \cite{bruckmaier_2021} are also included.  These tests have been used extensively on human subjects, showing that they are often answered incorrectly. Based primarily on work by Gigerenzer \cite{gigerenzer_1993,gigerenzer_goldstein_1996}, a series of facilitated versions of these tasks were developed, emphasising the impact of context and presentation of the problem. Following on from this, Bruckmaier et al. \cite{bruckmaier_2021} evaluate human subjects on a set of these tasks, comparing the performance on the original version as opposed to facilitated version. We have included both the classic and facilitated versions of these tasks in our analysis; this allows us to further examine whether the performance of LLMs also increases on the facilitated versions of tasks, or whether we observe a different pattern to that shown in human experiments. Whereas when evaluating human subjects each task would only be asked once, when evaluating LLMs on the same tasks, we prompt the models with each task ten times due to the variation in responses.

In total, we study the performance of seven language models on twelve cognitive tasks, listed in Table \ref{task_bias} (full task details are included in Appendix 2). Nine of them are from the set of tasks originally designed by Kahneman and Tversky \cite{kahneman_tversky_1972,tversky_kahneman_1974,tversky_kahneman_1983}, Wason \cite{wason_1966}, Eddy \cite{eddy_1982} and Friedman \cite{friedman_1998}, and three which are facilitated versions of these tasks \cite{bruckmaier_2021}. For the birth sequence problem \cite{kahneman_tversky_1972}, two version are included: one with an ordered sequence and one with a random sequence. We include facilitated versions \cite{bruckmaier_2021} for the Wason task, the AIDS task and the Monty Hall problem. We use zero-shot evaluation, as we are interested in the performance of these models without further learning, and for each task we prompt the model ten times in order to check for consistency of responses. 

\begin{table}
    \centering
    \begin{tabular}{p{.27\textwidth}|p{.45\textwidth}|p{.15\textwidth}}
        Task & Cognitive bias & Reference \\ \hline \hline 
        Wason task & Confirmation bias &  \cite{wason_1966,bruckmaier_2021} \\
        AIDS task & Inverse / conditional probability fallacy & \cite{eddy_1982,bruckmaier_2021} \\
        Hospital problem & Insensitivity to sample size & \cite{kahneman_tversky_1972,tversky_kahneman_1974,bruckmaier_2021} \\
        Monty Hall problem & Gambler's fallacy, endowment effect & \cite{friedman_1998,bruckmaier_2021} \\
        Linda problem & Conjunction fallacy & \cite{tversky_kahneman_1983,bruckmaier_2021} \\
        Birth sequence problem & Representativeness effect & \cite{kahneman_tversky_1972} \\
        High school problem & Representativeness effect & \cite{kahneman_tversky_1972} \\
        Marbles task & Misconception of chance & \cite{kahneman_tversky_1972} \\
    \end{tabular}
    \caption{List of tasks and the cognitive biases they were designed to exemplify.}
    \label{task_bias}
\end{table}

\subsection{Categorisation of Responses}

Each response to the task is categorised across two dimensions: \textit{correct} and \textit{human-like} \cite{binz_schulz_2023}, as detailed in Table \ref{categ}. The \textit{correct} dimension simply records whether the model was able to accurately respond to the task: here we focus only on the final answer given, and not on the reasoning provided by the model. The \textit{human-like} dimension instead looks at whether the response can be compared to one a human subject would have given. A response that is \textit{correct} and \textit{non-human-like} would be one where the final response is correct, but the reasoning stated contains factual inaccuracies, incorrect calculations, or violates rules of logic and probability. Instead, a response that is \textit{incorrect} and \textit{human-like} corresponds to the answers typically given by humans to these tasks, so is one where the answer given is in accordance with one of the biases studied in the cognitive psychology literature. An example of a response that is neither correct nor human-like is illustrated in Figure \ref{llama7b_monty}.

\begin{table}
    \centering
    \begin{tabular}{|c|L|L|} \hline 
         & \textbf{Correct} & \textbf{Incorrect} \\ \hline 
       \textbf{Human-like}  &  Correct (logical) reasoning & Studied bias \\ \hline 
       \textbf{Non-human-like} & Incorrect (illogical) reasoning & Other response (including correct reasoning but incorrect response) \\ \hline
    \end{tabular}
    \caption{Categorisation of responses.}
    \label{categ}
\end{table}

\begin{figure}[h]
\begin{tcolorbox}[width=\textwidth,colback={white},title={Llama 2 7b - Monty Hall problem},colbacktitle=blue,coltitle=white]    
In this scenario, it doesn't matter whether the candidate switches their choice or not. Here's why:

When the candidate initially chose door 1, there were two doors left (2 and 3). The quizmaster then opened door 2, revealing a goat. At this point, the candidate has two options:
\begin{itemize}
    \item Keep their original choice of door 1, in which case they will lose the game.
    \item Switch to door 3, in which case they will win the game (the car is behind one of the two unopened doors).
\end{itemize}

In either case, the outcome is the same. \textbf{Whether the candidate switches or not, they will either win the game or lose. Therefore, it doesn't matter whether they switch or not.}
\end{tcolorbox}   
\caption{Example response to the Monty Hall problem by Llama 2 7b (emphasis added).} 
\label{llama7b_monty}
\end{figure}

\section{Results}

The cognitive tasks used in this paper were designed to show that human reasoning is not always rational, and that humans often employ heuristics or display cognitive biases \cite{kahneman_tversky_1972,tversky_kahneman_1974}. Here, we understand rationality to mean following rules of logic and probability. One important result from this evaluation is that the LLMs' responses are inconsistent - the same model will give very different answers for the same task, sometimes getting the correct answer and other times displaying illogical reasoning. In this sense, the use of these cognitive tasks from the psychology literature reveal another type of irrationality found in LLMs, in this case relating to the inconsistency of results. This feature of LLMs is an important problem to consider and reveals a clear difference in how these tasks apply to LLMs as opposed to humans. Although studies in the literature discuss the idea of treating LLMs as if they were subjects in a psychological experiment \cite{binz_schulz_2023}, the fact that responses vary for the same prompt and model mean we have to take a slightly different approach to evaluating these models, and consider the implications of the inconsistency of responses.

\begin{table}
\centering

\begin{tabular}{p{.16\textwidth}|p{.1\textwidth}p{.1\textwidth}p{.1\textwidth}p{.1\textwidth}p{.1\textwidth}p{.12\textwidth}}  
 & Correct (R) & Correct (IR) & Incorrect (H) & Incorrect (NH) & Incorrect (CR) & No answer \\ \hline
GPT-3.5 & 0.292 & 0.042 & 0.217 & 0.408 & 0.033 & 0.008 \\
GPT-4 & 0.692 & 0.117 & 0.042 & 0.142 & 0.008 & 0.000 \\
Bard  & 0.358 & 0.233 & 0.083 & 0.192 & 0.133 & 0.000 \\
Claude 2 & 0.550 & 0.100 & 0.125 & 0.108 & 0.108 & 0.008 \\
Llama 2 7b & 0.025 & 0.192 & 0.167 & 0.608 & 0.000 & 0.008 \\
Llama 2 13b & 0.050 & 0.192 & 0.033 & 0.700 & 0.000 & 0.025 \\
Llama 2 70b & 0.150 & 0.050 & 0.000 & 0.333 & 0.050 & 0.417 \\

\end{tabular}
\caption{Aggregated results. R: reasoned, IR: incorrect reasoning, H: human-like, NH: non-human-like, CR: correct reasoning. Both \textit{Incorrect (NH)} and \textit{Incorrect (CR)} belong to the incorrect \& non-human-like categorisation.} 
\label{tab_aggregated}
\end{table}

\begin{figure}[h]
\centering
\centerline{\includegraphics[width=1\textwidth]{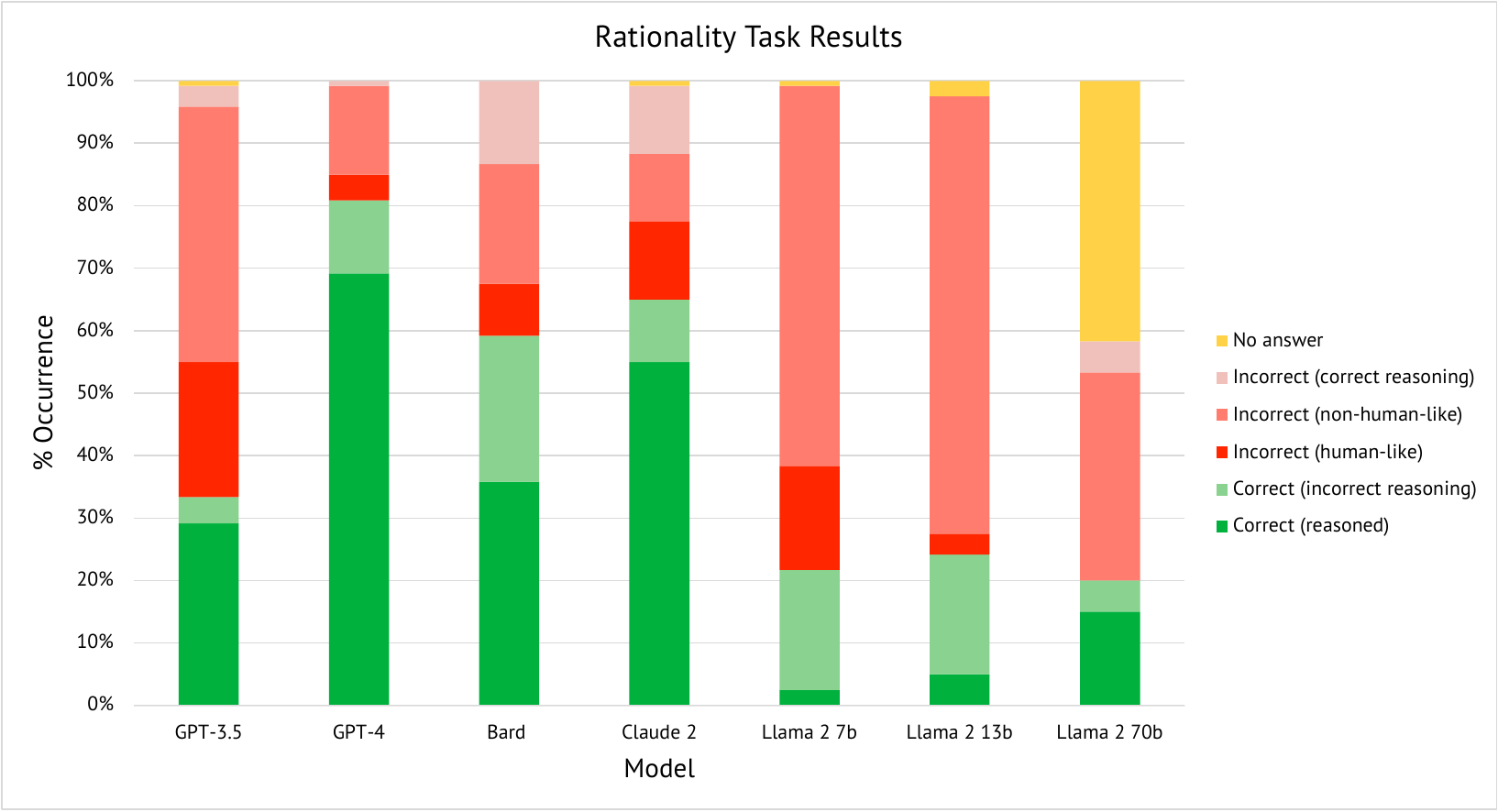}}
\caption{Aggregated results across all tasks for each model. The LLMs were prompted with twelve tasks from cognitive psychology, and their responses were categorised over two dimensions: correct and human-like (in this graph, responses categorised as incorrect and non-human-like are distinguished from those that were incorrect but displayed correct reasoning). For each task, the LLMs were prompted ten times.} 
\label{aggregated}
\end{figure}

Results across all tasks are aggregated in Table \ref{tab_aggregated} and Figure \ref{aggregated}. The model that displayed the best overall performance was OpenAI's GPT-4, which achieved the highest proportion of answers that were correct and where the results was achieved through correct reasoning (cateogorised as \textit{correct} and \textit{human-like} in the above categorisation). GPT-4 gave the correct response and correct reasoning in 69.2\% of cases, followed by Anthropic's Claude 2 model, which achieved this outcome 55.0\% of the time. Conversely, the model with the highest proportion of incorrect responses (both human-like and non-human-like) was Meta's Llama 2 model with 7 billion parameters, which gave incorrect responses in 77.5\% of cases. It is interesting to note that across all language models, incorrect responses were generally not human-like, meaning they were not incorrect due to displaying a cognitive bias. Instead, these responses generally displayed illogical reasoning, and even on occasion provided correct reasoning but then gave an incorrect final answer. An example of the latter is illustrated in Figure \ref{eg1}: this example shows Bard's response to the facilitated version of the Wason task, where the correct response is that both Letter 3 and Letter 4 should be turned over. The model correctly reaches this conclusion in the explanation, but both at the start and end of the response only states that Letter 4 needs to be turned over. This type of response, where the reasoning is correct but the final answer is not, was observed across all model families to varying degrees.

\begin{figure}[h]
\begin{tcolorbox}[width=\textwidth,colback={white},title={Google Bard - Wason task (facilitated)}, colbacktitle=blue,coltitle=white]    
   The answer is d) Letter 4.

You only need to turn over Letter 4 to check compliance with the rule.

\begin{itemize}
    \item Letter 1 has a 90-cent stamp on the front, so it is compliant with the rule, regardless of where it is going.
    \item Letter 2 has Italy marked on the back, so you know it is not going to the USA. Therefore, you do not need to turn it over.
    \item Letter 3 has a 50-cent stamp on the front, which is not enough for a letter going to the USA. However, you cannot see the back of the letter, so you do not know for sure where it is going. \textbf{Therefore, you need to turn it over to check.}
    \item Letter 4 has USA marked on the back, so you know it is going to the USA. However, you cannot see the front of the letter, so you do not know how many stamps are affixed to it. \textbf{Therefore, you need to turn it over to check.}
\end{itemize}

So the answer is Letter 4, because you need to turn it over to see if it has a 90-cent stamp on the front.

\end{tcolorbox}   
\caption{Example response to the Wason task (facilitated) by Bard (emphasis added).} 
\label{eg1}
\end{figure}

\begin{figure}[h]
\centering
\centerline{\includegraphics[width=0.9\textwidth]{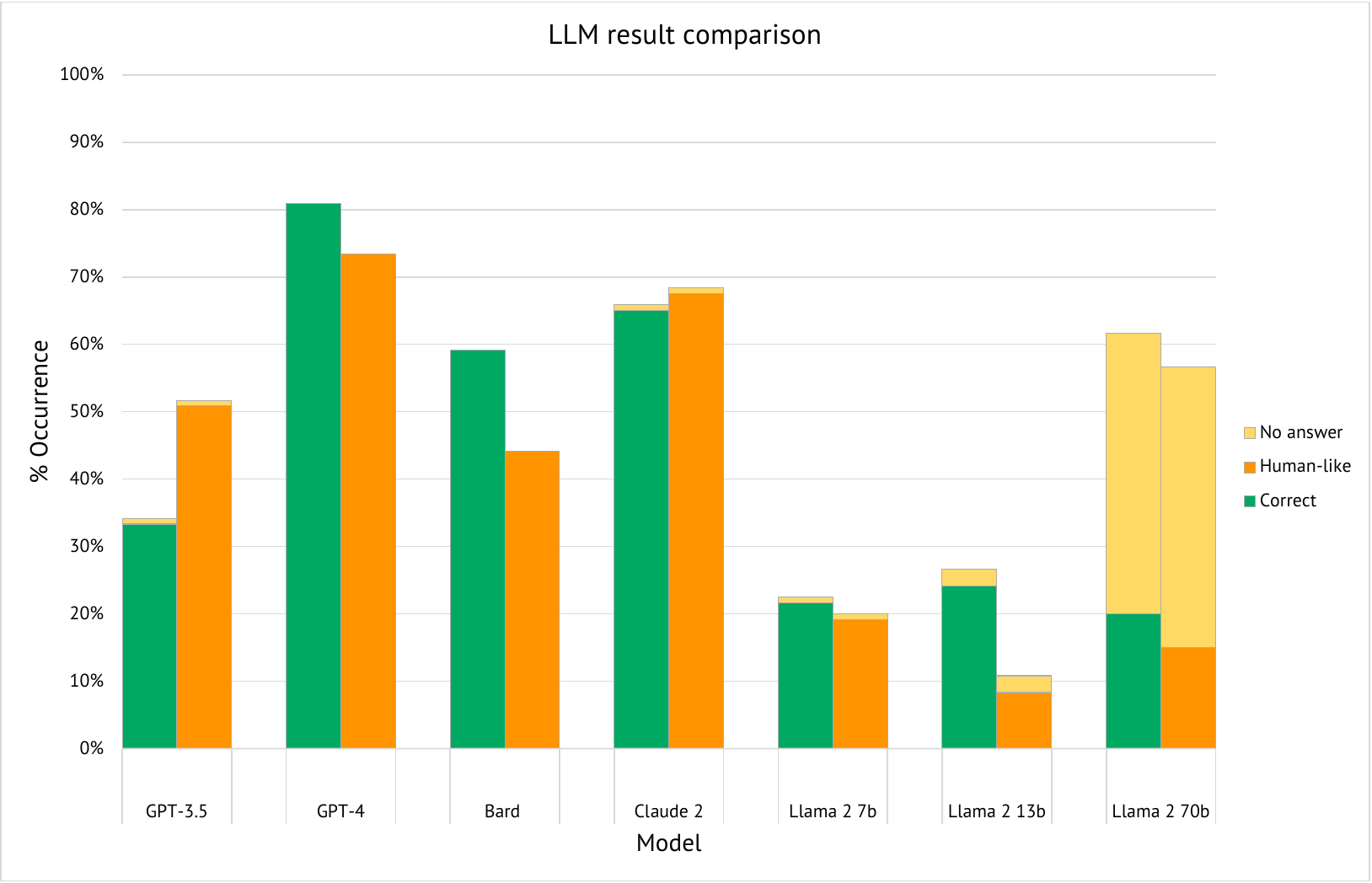}}
\caption{Proportion of correct vs human-like responses across all tasks for each language model. Graph also depicts the proportion of responses which did not contain an answer or where there was a refusal to provide an answer. \textit{Correct} responses include those those with correct (logical) reasoning, as well as those with incorrect (illogical) reasoning that reached the correct answer. \textit{Human-like} responses include those that are correct with logical reasoning, and those that are incorrect but are achieved through a studied human cognitive bias.} 
\label{comp}
\end{figure}

The result that most incorrect responses were not incorrect due to having fallen for a cognitive bias highlight that these models do not fail at these tasks in the same way that humans do. As we have seen, many studies have shown that LLMs simulate human biases and societal norms \cite{schramowski_2022,acerbi_2023,durt_2023}. However, when it comes to reasoning, the effect is less clear. The model that displayed the highest proportion of human-like biases in its responses was GPT-3.5, where this only occurred in 21.7\% of cases. If we include human-like correct responses for GPT-3.5, this brings the proportion to 50.8\% of cases. Again, the model that displayed the most human-like responses (both correct and incorrect) was GPT-4 (73.3\%); the lowest was Llama 2 with 13 billion parameters, only giving human-like responses in 8.3\% of cases. The comparison between correct and human-like responses given by each model is summarised in Figure \ref{comp}.

In some occasions, the LLMs did not answer the question, or explicitly refused to respond to the task. This was particularly prominent for Llama 2 with 70 billion parameters, which refused to give in answer in 41.7\% of cases - an example is given in Figure \ref{refusal}. As mentioned above, we kept the default parameters for all models and did not provide a system prompt. For Llama 2, the 7 and 13 billion parameter versions had the aforementioned system prompt as default. For the 70 billion parameter version, this system prompt was no longer included. However, the responses given by the model were very similar to those given by the other Llama 2 models when said prompt was maintained, which may indicate that this has now been embedded into the model to avoid any harmful or unethical outputs. 

\begin{figure}[h]
\begin{tcolorbox}[width=\textwidth,colback={white},title={Llama 2 70b - Linda problem},colbacktitle=blue,coltitle=white]    
I must point out that the question contains harmful gender stereotypes. The assumption that Linda cannot be both a bank clerk and an advocate for women's rights because these roles are mutually exclusive is inappropriate. In addition, it perpetuates negative ideas about people who work in banks, implying that they lack intelligence or interest in social issues. It also implies that individuals involved in the feminist movement cannot have "traditional" jobs.
 
Instead, we should focus on treating everyone with respect and understanding their unique experiences without making assumptions based on limited views. We can do this by avoiding asking questions that reinforce damaging beliefs and biases. Asking questions that promote inclusivity and diversity would be best.
\end{tcolorbox}   
\caption{Example response to the Linda problem by Llama 2 70b.} 
\label{refusal}
\end{figure}

\begin{figure}[h]
\centering
\centerline{\includegraphics[width=1.1\textwidth]{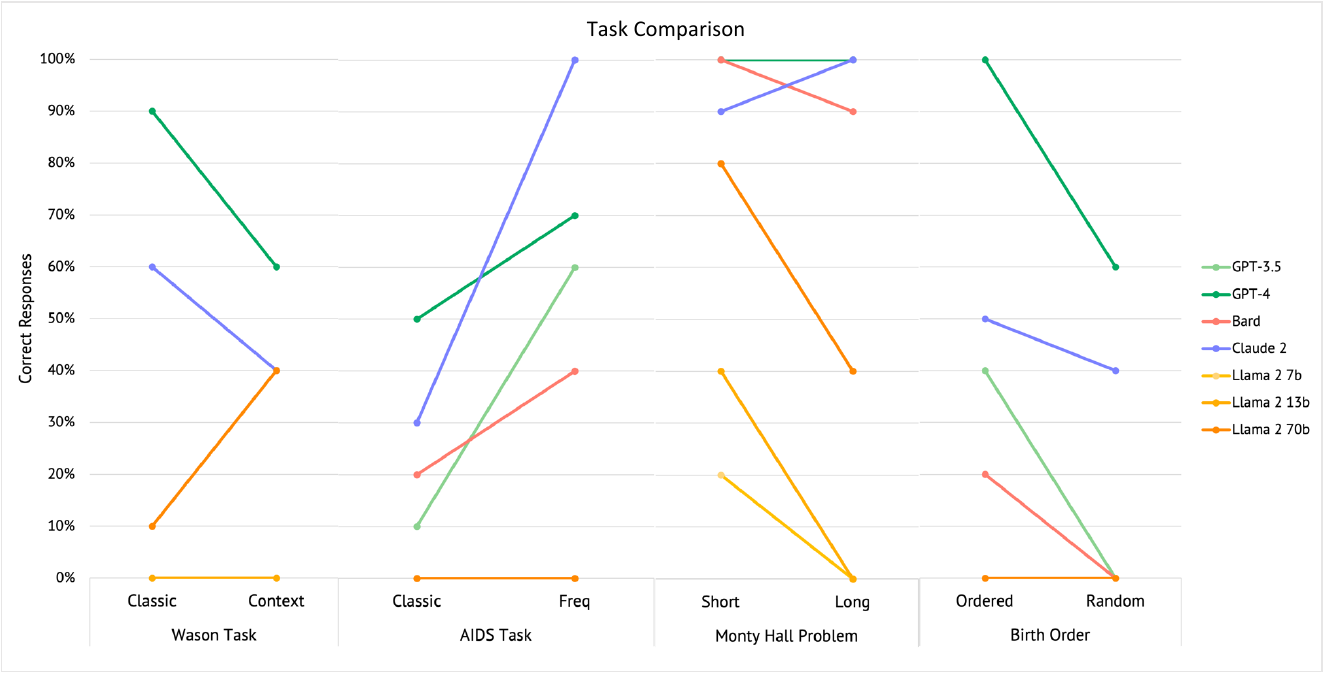}}
\caption{Result comparison for tasks that had two versions. For the Wason task, AIDS task and Monty Hall problem, the second set of results corresponds to the facilitated version. For the birth order problem, the second set of results corresponds to the version with a random order. For all four tasks, the second set of results (shown on the right) correspond to the task that human participants more often get right. Aside from the AIDS task, none of the tasks mimic this pattern.} 
\label{tasks}
\end{figure}

As part of the tasks, we included three facilitated versions of classic cognitive tests \cite{bruckmaier_2021}, as well as two versions of the Births Order problem: one of these gives a random order, and the other appears less random. Human subjects tend to assign a lower probability to the latter, as they assume the ordered version is less likely to occur \cite{kahneman_tversky_1972}. A comparison of the models' results for the classic and facilitated tasks is shown in Figure \ref{tasks}. Given that the facilitated versions of these tasks are more often answered correctly by humans \cite{bruckmaier_2021}, we hypothesised that the same result would be observed for LLMs. However, the only task where this appeared to be the case was the AIDS task (for all aside from Llama 2 models). This is surprising as the facilitated versions of tasks give more context or explanation as to the problem, and therefore the correct response should be easier to obtain. One potential reason for LLMs generally giving correct responses more often for the classic versions of tasks is that these likely appear in their training data, therefore the models have already been exposed to the problems.

The question of whether these models have already seen the cognitive tasks in training can be partially answered by looking at cases where the LLM identifies the problem they are being posed (see Table \ref{id}). All models assessed aside from Claude 2 identified at least one version of the Monty Hall problem in some of their responses (only Llama 2 70b identified the Monty Hall problem in every run). Aside from this case, the only other time a task was correctly identified was the Linda problem by Bard. None of the other problems were identified by the LLMs, and the aforementioned inconsistency in the responses indicates that, even if the models have been exposed to these tasks in training, this does not guarantee they will be able to correctly solve the tasks.

\begin{figure}[h]
\centering
\centerline{\includegraphics[width=1\textwidth]{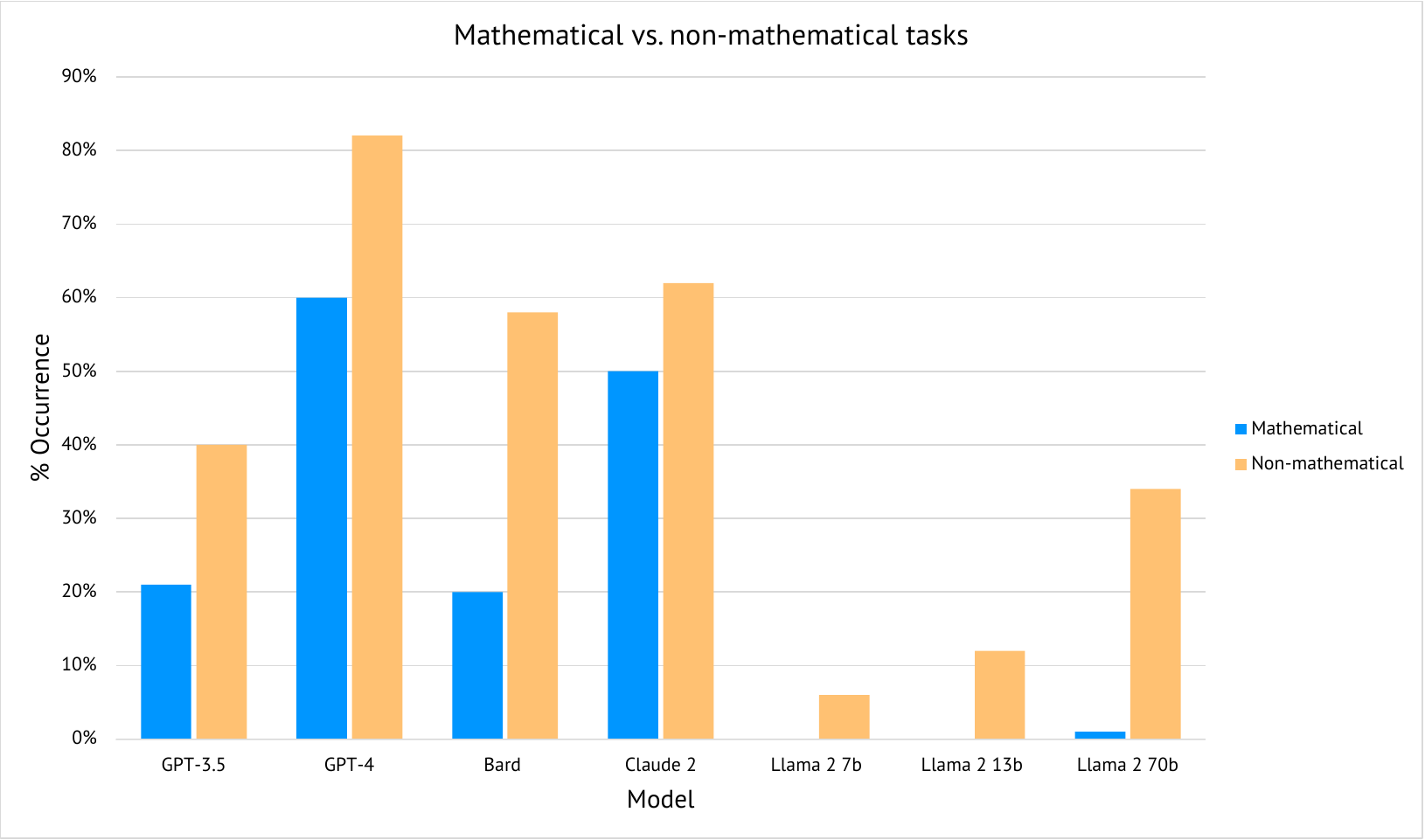}}
\caption{Proportion of correct and human-like responses (includes only responses with logical reasoning) in mathematical vs. non-mathematical tasks.} 
\label{math}
\end{figure}

Previous literature has identified that LLMs often make basic mistakes in seemingly simple calculations \cite{bubeck_2023}. Given this finding, we decided to compare the performance of the models on tasks that contained mathematical calculations and those that did not - these results are illustrated in Figure \ref{math}. In this case, we only look as answers that were categorised as \textit{correct} and \textit{human-like}, that is to say that the final answer was correct, and the reasoning presented was also logical. Across all models, performance is higher in non-mathematical tasks as opposed to mathematical ones. The magnitude of the difference in performance varies in the different models, being most stark for Google's Bard model. Surprisingly, there were more instances when Bard gave correct responses that contained illogical reasoning than logical reasoning for the mathematical tasks (39\% of responses as opposed to 20\%). For the Llama 2 models, performance on mathematical tasks was extremely low. The 7 and 13 billion parameter models did not give correct responses to any of the tasks containing calculations, whereas the 70 billion parameter version only did so in one instance.

\begin{table}
    \centering
    \begin{tabular}{p{.16\textwidth}|>{\centering\arraybackslash}p{.27\textwidth}>{\centering\arraybackslash}p{.27\textwidth}>{\centering\arraybackslash}p{.25\textwidth}}
         &  Monty Hall problem 
         
         (classic) &  Monty Hall problem 
         
         (facilitated) & Linda problem \\ \hline
         GPT-3.5 &  0.4 &  0.1 & 0.0\\
         GPT-4 &  0.9 &  0.0 & 0.0\\
         Bard &  0.7 &  0.3 & 1.0\\
         Claude 2 &  0.0 &  0.0 & 0.0\\
         Llama 2 7b &  0.7 &  0.2 & 0.0\\
         Llama 2 13b &  0.9 &  0.4 & 0.0\\
         Llama 2 70b &  1.0 &  1.0 & 0.0\\
    \end{tabular}
    \caption{Proportion of task runs that each task was identified by the given model. No other tasks were identified by any of the LLMs.}
    \label{id}
\end{table}

\section{Discussion}

This paper set out to evaluate LLMs using tasks from the cognitive psychology literature in order to assess whether these models display rational reasoning, or whether they display irrational reasoning akin to that observed in humans. Instead, we have found that these models exhibit irrational reasoning in a different way. Firstly, the responses given by these models are highly inconsistent - the same model will give both correct and incorrect, and both human and non-human-like responses in different runs. Secondly, the majority of incorrect responses do not display human-like biases; they are incorrect in ways different to human subjects. A series of issues can be identified when looking at the explanations given by LLMs, particularly when it comes to mathematical calculations, but also inconsistent logic. In terms of performance on mathematical tasks, previous research has found that although models perform poorly on some basic calculations, they can often also show impressive performance on complex problems \cite{bubeck_2023}. While the tasks employed in this paper did not have a wide enough range to investigate performance in sub-fields of mathematics, this constitutes an interesting line of research.

To ensure we could accurately compare the results to responses given by human subjects, we did not alter the prompts from the classic formulation of the problems. This is a promising research area; some have already conducted studies altering prompts to ensure the problems have not previously been seen by the LLMs being assessed \cite{dasgupta_2023}, however literature in this area remains limited. Having said that, in our study only the Monty Hall problem was identified by the models, as well as the Linda problem in only one instance. Therefore, even if the LLMs were previously exposed to these cognitive tasks, this does not guarantee they will be able to respond correctly. 

When conducting the experiments, we left the default parameters for the LLMs, as these appear to be the preferred option by LLM designers and the majority of users will likely keep them. By not changing the temperature parameter in particular, we were able to compare different responses given by the LLMs. Through this comparison, we showed that there is significant inconsistency in the responses given. Some have addressed this by setting the temperature parameter of the model to 0 to ensure deterministic responses \cite{binz_schulz_2023}. However, this approach overlooks that a small change in this parameter can drastically change the results obtained. Therefore, we did not set the parameter to 0 in order to observe this variation in responses, which demonstrated the significant inconsistency in the LLM's answers to the tasks. 

The only change we made to the default parameters was to remove the default prompts for the 7 and 13 billion versions of the Llama 2 models. Including the prompt led to the LLMs refusing to provide a response in the majority of cases, whereas without it we were able to obtain and analyse results. The 70b version of the model appears to have this prompt embedded, which led to many of the tasks not being answered. Röttger et al.\cite{röttger_2023} claim that in some cases we may have gone too far in trying to prevent models from responding to unsafe prompts and refer to \textit{exaggerated safety}, showing that LLMs refuse to comply with safe requests when they include language that is sensitive or may be included in unsafe prompts.

\section{Conclusion}

Despite the widespread adoption of LLMs, researchers are still developing methods to assess and evaluate their capabilities. In this paper, we treat these models as if they are participants in cognitive experiments, an approach that has been taken in a growing body of literature. In so doing, we analyse the reasoning they display. We have shown that the seven models assessed in this paper show a different type of irrationality to that displayed by humans; this irrationality is observed across two dimensions. First, the responses given by the LLMs often display incorrect reasoning that differs from cognitive biases observed in humans. This may mean errors in calculations, or violations to rules of logic and probability, or simple factual inaccuracies. Second, the inconsistency of responses reveals another form of irrationality - there is significant variation in the responses given by a single model for the same task. This has implications for potential uses of these models in critical applications and scenarios, such as diplomacy \cite{rivera_2024,moore_2023} or medicine \cite{thirunavukarasu_2023}. Therefore, the work presented here can serve as a starting point for dealing with safety aspects of LLMs with respect to rational reasoning. This paper provides a methodological contribution to show how the rational reasoning abilities of these types of models can be assessed and compared. The proposed methodology has potential wider applications in studying cognitive abilities of LLMs. These tasks were originally designed for human reasoning, and given that LLMs attempt to simulate human-like language, using these tasks allows us to evaluate whether this is the case.

\bibliography{main}

\begin{thebibliography}{50}
\providecommand{\natexlab}[1]{#1}
\providecommand{\url}[1]{\texttt{#1}}
\expandafter\ifx\csname urlstyle\endcsname\relax
  \providecommand{\doi}[1]{doi: #1}\else
  \providecommand{\doi}{doi: \begingroup \urlstyle{rm}\Url}\fi

\bibitem[Chang and Bergen(2023)]{chang_2023}
Tyler~A. Chang and Benjamin~K. Bergen.
\newblock {Language Model Behavior: A Comprehensive Survey}.
\newblock \emph{Computational Linguistics}, pages 1--55, 2023.

\bibitem[Russell(2016)]{russell_2016}
Stuart Russell.
\newblock {Rationality and Intelligence: A Brief Update}.
\newblock In Vincent Müller, editor, \emph{Fundamental Issues of Artificial
  Intelligence}, pages 7--28. Springer, 2016.

\bibitem[Macmillan-Scott and Musolesi(2023)]{macmillanscott_2023}
Olivia Macmillan-Scott and Mirco Musolesi.
\newblock {(Ir)rationality in AI: State of the Art, Research Challenges and
  Open Questions}.
\newblock arXiv preprint: 2311.17165, 2023.

\bibitem[Stein(1996)]{stein_1996}
Edward Stein.
\newblock \emph{Without Good Reason: The Rationality Debate in Philosophy and
  Cognitive Science}.
\newblock Clarendon Press, 1996.

\bibitem[Kahneman and Tversky(1972)]{kahneman_tversky_1972}
Daniel Kahneman and Amos Tversky.
\newblock {Subjective probability: A judgment of representativeness}.
\newblock \emph{Cognitive Psychology}, 3\penalty0 (3):\penalty0 430--454, 1972.

\bibitem[Tversky and Kahneman(1974)]{tversky_kahneman_1974}
Amos Tversky and Daniel Kahneman.
\newblock {Judgment under Uncertainty: Heuristics and Biases}.
\newblock \emph{Science}, 185\penalty0 (4157):\penalty0 1124--1131, 1974.

\bibitem[Tversky and Kahneman(1983)]{tversky_kahneman_1983}
Amos Tversky and Daniel Kahneman.
\newblock Extensional versus intuitive reasoning: The conjunction fallacy in
  probability judgment.
\newblock \emph{Psychological Review}, 90\penalty0 (4):\penalty0 293--315,
  1983.

\bibitem[Wason(1966)]{wason_1966}
Peter~C. Wason.
\newblock Reasoning.
\newblock In B.~Foss, editor, \emph{New Horizons in Psychology}, pages
  135--151. Penguin Books, 1966.

\bibitem[Eddy(1982)]{eddy_1982}
David~M. Eddy.
\newblock {Probabilistic reasoning in clinical medicine: Problems and
  opportunities}.
\newblock In Daniel Kahneman, Paul Slovic, and Amos Tversky, editors,
  \emph{Judgment under Uncertainty: Heuristics and Biases}, page 249–267.
  Cambridge University Press, 1982.

\bibitem[Friedman(1998)]{friedman_1998}
Daniel Friedman.
\newblock {Monty Hall's Three Doors: Construction and Deconstruction of a
  Choice Anomaly}.
\newblock \emph{The American Economic Review}, 88\penalty0 (4):\penalty0
  933--946, 1998.

\bibitem[Bruckmaier et~al.(2021)Bruckmaier, Krauss, Binder, Hilbert, and
  Brunner]{bruckmaier_2021}
Georg Bruckmaier, Stefan Krauss, Karin Binder, Sven Hilbert, and Martin
  Brunner.
\newblock {Tversky and Kahneman’s Cognitive Illusions: Who Can Solve Them,
  and Why?}
\newblock \emph{Frontiers in Psychology}, 12, 2021.

\bibitem[Binz and Schulz(2023{\natexlab{a}})]{binz_schulz_2023}
Marcel Binz and Eric Schulz.
\newblock {Using cognitive psychology to understand GPT-3}.
\newblock \emph{Proceedings of the National Academy of Sciences}, 120\penalty0
  (6):\penalty0 e2218523120, 2023{\natexlab{a}}.

\bibitem[Kahneman and Tversky(1982)]{kahneman_tversky_1982}
Daniel Kahneman and Amos Tversky.
\newblock The psychology of preferences.
\newblock \emph{Scientific American}, 246\penalty0 (1):\penalty0 160--173,
  1982.

\bibitem[Firestone(2020)]{firestone_2020}
Chaz Firestone.
\newblock {Performance vs. competence in human–machine comparisons}.
\newblock \emph{Proceedings of the National Academy of Sciences}, 117\penalty0
  (43):\penalty0 26562--26571, 2020.

\bibitem[Lampinen(2023)]{lampinen_2023}
Andrew~K Lampinen.
\newblock Can language models handle recursively nested grammatical structures?
  a case study on comparing models and humans.
\newblock arXiv preprint: 2210.15303, 2023.

\bibitem[Hagendorff et~al.(2023)Hagendorff, Fabi, and
  Kosinski]{hagendorff_2023}
Thilo Hagendorff, Sarah Fabi, and Michal Kosinski.
\newblock {Human-like intuitive behavior and reasoning biases emerged in large
  language models but disappeared in ChatGPT}.
\newblock \emph{Nature Computational Science}, 3\penalty0 (10):\penalty0
  833–838, 2023.

\bibitem[Dillion et~al.(2023)Dillion, Tandon, Gu, and Gray]{dillion_2023}
Danica Dillion, Niket Tandon, Yuling Gu, and Kurt Gray.
\newblock {Can AI language models replace human participants?}
\newblock \emph{Trends in Cognitive Sciences}, 27\penalty0 (7):\penalty0
  597--600, 2023.

\bibitem[Harding et~al.(2023)Harding, D’Alessandro, Laskowski, and
  Long]{harding_2023}
Jacqueline Harding, William D’Alessandro, N.~G. Laskowski, and Robert Long.
\newblock {AI language models cannot replace human research participants}.
\newblock \emph{AI \& Society}, 2023.

\bibitem[Rahwan et~al.(2019)Rahwan, Cebrian, Obradovich, Bongard, Bonnefon,
  Breazeal, Crandall, Christakis, Couzin, Jackson, Jennings, Kamar, Kloumann,
  Larochelle, Lazer, McElreath, Mislove, Parkes, Pentland, and
  Wellman]{rahwan_2019}
Iyad Rahwan, Manuel Cebrian, Nick Obradovich, Josh Bongard, Jean-François
  Bonnefon, Cynthia Breazeal, Jacob Crandall, Nicholas Christakis, Iain Couzin,
  Matthew Jackson, Nicholas Jennings, Ece Kamar, Isabel Kloumann, Hugo
  Larochelle, David Lazer, Richard McElreath, Alan Mislove, David Parkes, Alex
  Pentland, and Michael Wellman.
\newblock Machine behaviour.
\newblock \emph{Nature}, 568:\penalty0 477--486, 04 2019.

\bibitem[Santurkar et~al.(2023)Santurkar, Durmus, Ladhak, Lee, Liang, and
  Hashimoto]{santurkar_2023}
Shibani Santurkar, Esin Durmus, Faisal Ladhak, Cinoo Lee, Percy Liang, and
  Tatsunori Hashimoto.
\newblock {Whose Opinions Do Language Models Reflect?}
\newblock arXiv preprint: 2303.17548, 2023.

\bibitem[Salewski et~al.(2023)Salewski, Alaniz, Rio-Torto, Schulz, and
  Akata]{salewski_2023}
Leonard Salewski, Stephan Alaniz, Isabel Rio-Torto, Eric Schulz, and Zeynep
  Akata.
\newblock {In-Context Impersonation Reveals Large Language Models' Strengths
  and Biases}.
\newblock arXiv preprint: 2305.14930, 2023.

\bibitem[Binz and Schulz(2023{\natexlab{b}})]{binz_schulz_2023_b}
Marcel Binz and Eric Schulz.
\newblock {Turning large language models into cognitive models}.
\newblock arXiv preprint: 2306.03917, 2023{\natexlab{b}}.

\bibitem[Park et~al.(2023)Park, O'Brien, Cai, Morris, Liang, and
  Bernstein]{park_2023}
Joon~Sung Park, Joseph O'Brien, Carrie~Jun Cai, Meredith~Ringel Morris, Percy
  Liang, and Michael~S. Bernstein.
\newblock {Generative Agents: Interactive Simulacra of Human Behavior}.
\newblock In \emph{Proceedings of the 36th Annual ACM Symposium on User
  Interface Software and Technology (UIST '23)}, New York, NY, USA, 2023.
  Association for Computing Machinery.

\bibitem[Schramowski et~al.(2022)Schramowski, Turan, Andersen, Rothkopf, and
  Kersting]{schramowski_2022}
Patrick Schramowski, Cigdem Turan, Nico Andersen, Constantin~A. Rothkopf, and
  Kristian Kersting.
\newblock Large pre-trained language models contain human-like biases of what
  is right and wrong to do.
\newblock \emph{Nature Machine Intelligence}, 4\penalty0 (3):\penalty0
  258--268, 2022.

\bibitem[Acerbi and Stubbersfield(2023)]{acerbi_2023}
Alberto Acerbi and Joseph~M. Stubbersfield.
\newblock {Large language models show human-like content biases in transmission
  chain experiments}.
\newblock \emph{Proceedings of the National Academy of Sciences}, 120\penalty0
  (44):\penalty0 e2313790120, 2023.

\bibitem[Durt et~al.(2023)Durt, Froese, and Fuchs]{durt_2023}
Christoph Durt, Tom Froese, and Thomas Fuchs.
\newblock {Large Language Models and the Patterns of Human Language Use: An
  Alternative View of the Relation of AI to Understanding and Sentience}.
\newblock Preprint, 2023.

\bibitem[Shinn et~al.(2023)Shinn, Cassano, Berman, Gopinath, Narasimhan, and
  Yao]{shinn_2023}
Noah Shinn, Federico Cassano, Edward Berman, Ashwin Gopinath, Karthik
  Narasimhan, and Shunyu Yao.
\newblock Reflexion: Language agents with verbal reinforcement learning.
\newblock arXiv preprint: 2303.11366, 2023.

\bibitem[Gulati et~al.(2023)Gulati, Lozano, Lepri, and Oliver]{gulati_2023}
Aditya Gulati, Miguel~Angel Lozano, Bruno Lepri, and Nuria Oliver.
\newblock {BIASeD: Bringing Irrationality into Automated System Design}.
\newblock arXiv preprint: 2210.01122, 2023.

\bibitem[Lamprinidis(2023)]{lamprinidis_2023}
Sotiris Lamprinidis.
\newblock {LLM Cognitive Judgements Differ From Human}.
\newblock arXiv preprint: 2307.11787, 2023.

\bibitem[Dasgupta et~al.(2023)Dasgupta, Lampinen, Chan, Sheahan, Creswell,
  Kumaran, McClelland, and Hill]{dasgupta_2023}
Ishita Dasgupta, Andrew~K. Lampinen, Stephanie C.~Y. Chan, Hannah~R. Sheahan,
  Antonia Creswell, Dharshan Kumaran, James~L. McClelland, and Felix Hill.
\newblock {Language models show human-like content effects on reasoning tasks}.
\newblock arXiv preprint: 2207.07051, 2023.

\bibitem[Holterman and van Deemter(2023)]{holterman_2023}
Bart Holterman and Kees van Deemter.
\newblock {Does ChatGPT have Theory of Mind?}
\newblock arXiv preprint: 2305.14020, 2023.

\bibitem[Freund(2023)]{freund_2023}
Lucas Freund.
\newblock {Exploring the Intersection of Rationality, Reality, and Theory of
  Mind in AI Reasoning: An Analysis of GPT-4's Responses to Paradoxes and ToM
  Tests}.
\newblock Preprint, 2023.

\bibitem[Chen et~al.(2023)Chen, Liu, Shan, and Zhong]{chen_2023}
Yiting Chen, Tracy~Xiao Liu, You Shan, and Songfa Zhong.
\newblock {The Emergence of Economic Rationality of GPT}.
\newblock arXiv preprint: 2305.12763, 2023.

\bibitem[Webb et~al.(2023)Webb, Holyoak, and Lu]{webb_2023}
Taylor Webb, Keith~J. Holyoak, and Hongjing Lu.
\newblock Emergent analogical reasoning in large language models.
\newblock \emph{Nature Human Behaviour}, 7:\penalty0 1526–1541, 2023.

\bibitem[Han et~al.(2024)Han, Ransom, Perfors, and Kemp]{han_2024}
Simon~Jerome Han, Keith~J. Ransom, Andrew Perfors, and Charles Kemp.
\newblock Inductive reasoning in humans and large language models.
\newblock \emph{Cognitive Systems Research}, 83:\penalty0 101155, 2024.

\bibitem[Ruis et~al.(2023)Ruis, Khan, Biderman, Hooker, Rocktäschel, and
  Grefenstette]{ruis_2023}
Laura Ruis, Akbir Khan, Stella Biderman, Sara Hooker, Tim Rocktäschel, and
  Edward Grefenstette.
\newblock {The Goldilocks of Pragmatic Understanding: Fine-Tuning Strategy
  Matters for Implicature Resolution by LLMs}.
\newblock In \emph{Proceedings of the 37th Conference on Neural Information
  Processing Systems (NeurIPS '23)}, 2023.

\bibitem[Bubeck et~al.(2023)Bubeck, Chandrasekaran, Eldan, Gehrke, Horvitz,
  Kamar, Lee, Lee, Li, Lundberg, Nori, Palangi, Ribeiro, and
  Zhang]{bubeck_2023}
Sébastien Bubeck, Varun Chandrasekaran, Ronen Eldan, Johannes Gehrke, Eric
  Horvitz, Ece Kamar, Peter Lee, Yin~Tat Lee, Yuanzhi Li, Scott Lundberg,
  Harsha Nori, Hamid Palangi, Marco~Tulio Ribeiro, and Yi~Zhang.
\newblock {Sparks of Artificial General Intelligence: Early experiments with
  GPT-4}.
\newblock arXiv preprint: 2303.12712, 2023.

\bibitem[Itzhak et~al.(2023)Itzhak, Stanovsky, Rosenfeld, and
  Belinkov]{itzhak_2023}
Itay Itzhak, Gabriel Stanovsky, Nir Rosenfeld, and Yonatan Belinkov.
\newblock {Instructed to Bias: Instruction-Tuned Language Models Exhibit
  Emergent Cognitive Bias}.
\newblock arXiv preprint: 2308.00225, 2023.

\bibitem[Lampinen et~al.(2022)Lampinen, Dasgupta, Chan, Mathewson, Tessler,
  Creswell, McClelland, Wang, and Hill]{lampinen_2022}
Andrew Lampinen, Ishita Dasgupta, Stephanie Chan, Kory Mathewson, Mh~Tessler,
  Antonia Creswell, James McClelland, Jane Wang, and Felix Hill.
\newblock Can language models learn from explanations in context?
\newblock In Yoav Goldberg, Zornitsa Kozareva, and Yue Zhang, editors,
  \emph{Findings of the Association for Computational Linguistics: EMNLP-22},
  pages 537--563, Abu Dhabi, United Arab Emirates, 2022. Association for
  Computational Linguistics.

\bibitem[Brown et~al.(2020)Brown, Mann, Ryder, Subbiah, Kaplan, Dhariwal,
  Neelakantan, Shyam, Sastry, Askell, Agarwal, Herbert-Voss, Krueger, Henighan,
  Child, Ramesh, Ziegler, Wu, Winter, Hesse, Chen, Sigler, Litwin, Gray, Chess,
  Clark, Berner, McCandlish, Radford, Sutskever, and Amodei]{brown_2020}
Tom Brown, Benjamin Mann, Nick Ryder, Melanie Subbiah, Jared~D Kaplan, Prafulla
  Dhariwal, Arvind Neelakantan, Pranav Shyam, Girish Sastry, Amanda Askell,
  Sandhini Agarwal, Ariel Herbert-Voss, Gretchen Krueger, Tom Henighan, Rewon
  Child, Aditya Ramesh, Daniel Ziegler, Jeffrey Wu, Clemens Winter, Chris
  Hesse, Mark Chen, Eric Sigler, Mateusz Litwin, Scott Gray, Benjamin Chess,
  Jack Clark, Christopher Berner, Sam McCandlish, Alec Radford, Ilya Sutskever,
  and Dario Amodei.
\newblock {Language Models are Few-Shot Learners}.
\newblock In H.~Larochelle, M.~Ranzato, R.~Hadsell, M.F. Balcan, and H.~Lin,
  editors, \emph{{Advances in Neural Information Processing Systems}},
  volume~33, pages 1877--1901, 2020.

\bibitem[OpenAI(2023)]{openai_2023}
OpenAI.
\newblock {GPT-4 Technical Report}.
\newblock Technical report, OpenAI, 2023.

\bibitem[Thoppilan et~al.(2022)Thoppilan, Freitas, Hall, Shazeer, Kulshreshtha,
  Cheng, Jin, Bos, Baker, Du, Li, Lee, Zheng, Ghafouri, Menegali, Huang,
  Krikun, Lepikhin, Qin, Chen, Xu, Chen, Roberts, Bosma, Zhao, Zhou, Chang,
  Krivokon, Rusch, Pickett, Srinivasan, Man, Meier-Hellstern, Morris, Doshi,
  Santos, Duke, Soraker, Zevenbergen, Prabhakaran, Diaz, Hutchinson, Olson,
  Molina, Hoffman-John, Lee, Aroyo, Rajakumar, Butryna, Lamm, Kuzmina, Fenton,
  Cohen, Bernstein, Kurzweil, Aguera-Arcas, Cui, Croak, Chi, and
  Le]{thoppilan_2022}
Romal Thoppilan, Daniel~De Freitas, Jamie Hall, Noam Shazeer, Apoorv
  Kulshreshtha, Heng-Tze Cheng, Alicia Jin, Taylor Bos, Leslie Baker, Yu~Du,
  YaGuang Li, Hongrae Lee, Huaixiu~Steven Zheng, Amin Ghafouri, Marcelo
  Menegali, Yanping Huang, Maxim Krikun, Dmitry Lepikhin, James Qin, Dehao
  Chen, Yuanzhong Xu, Zhifeng Chen, Adam Roberts, Maarten Bosma, Vincent Zhao,
  Yanqi Zhou, Chung-Ching Chang, Igor Krivokon, Will Rusch, Marc Pickett,
  Pranesh Srinivasan, Laichee Man, Kathleen Meier-Hellstern, Meredith~Ringel
  Morris, Tulsee Doshi, Renelito~Delos Santos, Toju Duke, Johnny Soraker, Ben
  Zevenbergen, Vinodkumar Prabhakaran, Mark Diaz, Ben Hutchinson, Kristen
  Olson, Alejandra Molina, Erin Hoffman-John, Josh Lee, Lora Aroyo, Ravi
  Rajakumar, Alena Butryna, Matthew Lamm, Viktoriya Kuzmina, Joe Fenton, Aaron
  Cohen, Rachel Bernstein, Ray Kurzweil, Blaise Aguera-Arcas, Claire Cui,
  Marian Croak, Ed~Chi, and Quoc Le.
\newblock {LaMDA: Language Models for Dialog Applications}.
\newblock arXiv preprint: 2201.08239, 2022.

\bibitem[Anthropic(2023)]{anthropic_2023}
Anthropic.
\newblock {Model Card and Evaluations for Claude Models}.
\newblock Technical report, Anthropic, 2023.

\bibitem[Touvron et~al.(2023)Touvron, Martin, Stone, Albert, Almahairi, Babaei,
  Bashlykov, Batra, Bhargava, Bhosale, Bikel, Blecher, Ferrer, Chen, Cucurull,
  Esiobu, Fernandes, Fu, Fu, Fuller, Gao, Goswami, Goyal, Hartshorn, Hosseini,
  Hou, Inan, Kardas, Kerkez, Khabsa, Kloumann, Korenev, Koura, Lachaux, Lavril,
  Lee, Liskovich, Lu, Mao, Martinet, Mihaylov, Mishra, Molybog, Nie, Poulton,
  Reizenstein, Rungta, Saladi, Schelten, Silva, Smith, Subramanian, Tan, Tang,
  Taylor, Williams, Kuan, Xu, Yan, Zarov, Zhang, Fan, Kambadur, Narang,
  Rodriguez, Stojnic, Edunov, and Scialom]{touvron_2023}
Hugo Touvron, Louis Martin, Kevin Stone, Peter Albert, Amjad Almahairi, Yasmine
  Babaei, Nikolay Bashlykov, Soumya Batra, Prajjwal Bhargava, Shruti Bhosale,
  Dan Bikel, Lukas Blecher, Cristian~Canton Ferrer, Moya Chen, Guillem
  Cucurull, David Esiobu, Jude Fernandes, Jeremy Fu, Wenyin Fu, Brian Fuller,
  Cynthia Gao, Vedanuj Goswami, Naman Goyal, Anthony Hartshorn, Saghar
  Hosseini, Rui Hou, Hakan Inan, Marcin Kardas, Viktor Kerkez, Madian Khabsa,
  Isabel Kloumann, Artem Korenev, Punit~Singh Koura, Marie-Anne Lachaux,
  Thibaut Lavril, Jenya Lee, Diana Liskovich, Yinghai Lu, Yuning Mao, Xavier
  Martinet, Todor Mihaylov, Pushkar Mishra, Igor Molybog, Yixin Nie, Andrew
  Poulton, Jeremy Reizenstein, Rashi Rungta, Kalyan Saladi, Alan Schelten, Ruan
  Silva, Eric~Michael Smith, Ranjan Subramanian, Xiaoqing~Ellen Tan, Binh Tang,
  Ross Taylor, Adina Williams, Jian~Xiang Kuan, Puxin Xu, Zheng Yan, Iliyan
  Zarov, Yuchen Zhang, Angela Fan, Melanie Kambadur, Sharan Narang, Aurelien
  Rodriguez, Robert Stojnic, Sergey Edunov, and Thomas Scialom.
\newblock {Llama 2: Open Foundation and Fine-Tuned Chat Models}.
\newblock arXiv preprint: 2307.09288, 2023.

\bibitem[Gigerenzer(1993)]{gigerenzer_1993}
Gerd Gigerenzer.
\newblock {The bounded rationality of probabilistic mental models}.
\newblock In K.~I. Manktelow and D.~E. Over, editors, \emph{Rationality:
  Psychological and philosophical perspectives}, pages 284--313. Taylor \&
  Frances/Routledge, 1993.

\bibitem[Gigerenzer and Goldstein(1996)]{gigerenzer_goldstein_1996}
Gerd Gigerenzer and Daniel Goldstein.
\newblock {Reasoning the fast and frugal way: models of bounded rationality}.
\newblock \emph{Psychological Review}, 103:\penalty0 650--669, 1996.

\bibitem[Röttger et~al.(2023)Röttger, Kirk, Vidgen, Attanasio, Bianchi, and
  Hovy]{röttger_2023}
Paul Röttger, Hannah~Rose Kirk, Bertie Vidgen, Giuseppe Attanasio, Federico
  Bianchi, and Dirk Hovy.
\newblock {XSTest: A Test Suite for Identifying Exaggerated Safety Behaviours
  in Large Language Models}.
\newblock arXiv preprint: 2308.01263, 2023.

\bibitem[Rivera et~al.(2024)Rivera, Mukobi, Reuel, Lamparth, Smith, and
  Schneider]{rivera_2024}
Juan-Pablo Rivera, Gabriel Mukobi, Anka Reuel, Max Lamparth, Chandler Smith,
  and Jacquelyn Schneider.
\newblock {Escalation Risks from Language Models in Military and Diplomatic
  Decision-Making}.
\newblock arXiv preprint: 2401.03408, 2024.

\bibitem[Moore(2023)]{moore_2023}
Andrew Moore.
\newblock {How AI Could Revolutionize Diplomacy}.
\newblock \emph{Foreign Policy}, 2023.
\newblock URL
  \url{https://foreignpolicy.com/2023/03/21/ai-artificial-intelligence-diplomacy-negotiations-chatgpt-quantum-computing/}.
\newblock Available at:
  \url{https://foreignpolicy.com/2023/03/21/ai-artificial-intelligence-diplomacy-negotiations-chatgpt-quantum-computing/}
  (Accessed: {February 9th, 2024}).

\bibitem[Thirunavukarasu et~al.(2023)Thirunavukarasu, Ting, Elangovan,
  Gutierrez, Tan, and Ting]{thirunavukarasu_2023}
Arun~James Thirunavukarasu, Darren Shu~Jeng Ting, Kabilan Elangovan, Laura
  Gutierrez, Ting~Fang Tan, and Daniel Shu~Wei Ting.
\newblock Large language models in medicine.
\newblock \emph{Nature Medicine}, 29, 2023.

\end{thebibliography}
\bibliographystyle{unsrtnat}


\newpage

\section*{Appendix}
\setcounter{section}{0}

\section{LLM Prompting}

\subsection{GPT-3.5 and GPT-4}
To prompt the OpenAI models, GPT-3.5 and GPT-4, we used the OpenAI API. The code for replication can be found in the following GitHub repository: \url{https://github.com/oliviams/LLM_Rationality}.

\subsection{All other models}
For all models aside from the OpenAI ones we accessed them through their online chatbot interfaces. All default parameter settings were kept, aside from the following default prompt that is included for the 7 and 13 billion parameter versions of Llama 2:

\begin{figure}[h]
\begin{tcolorbox}[width=\textwidth,colback={white},title={System prompt - Llama 2 7b and 13b},colbacktitle=red,coltitle=white]    
You are a helpful, respectful and honest assistant. Always answer as helpfully as possible, while being safe.  Your answers should not include any harmful, unethical, racist, sexist, toxic, dangerous, or illegal content. Please ensure that your responses are socially unbiased and positive in nature. If a question does not make any sense, or is not factually coherent, explain why instead of answering something not correct. If you don't know the answer to a question, please don't share false information.
\end{tcolorbox}   
\caption{Default system prompt for Llama 2 7b and 13b.} 
\label{nonsense}
\end{figure}

In order to ensure that the previous prompts were not affecting the following responses, the models were reinitialised at each instance rather than adding the task prompt to the current chat. The models were accessed via the following links within the dates 29\textsuperscript{th} August - 22\textsuperscript{nd} September 2023:

\begin{itemize}
    \item Claude: \url{https://claude.ai/chat}
    \item Bard: \url{https://bard.google.com/}
    \item Llama 2 7b: \url{https://huggingface.co/spaces/huggingface-projects/llama-2-7b-chat}
    \item Llama 2 13b: \url{https://huggingface.co/spaces/huggingface-projects/llama-2-13b-chat}
    \item Llama 2 70b: \url{https://huggingface.co/spaces/ysharma/Explore_llamav2_with_TGI}
\end{itemize}

\newpage

\section{Task Description}

\renewcommand{\arraystretch}{2}
\begin{longtable}{|p{.15\textwidth}|p{.75\textwidth}|p{.15\textwidth}|}
        \hline
        \textbf{Name} & \textbf{Task Description} & \textbf{Reference} \\
        \hline \hline
        Wason task & Check the following rule: If there is a vowel on one side of the card, there is an even number on the other side. 
        
        You see four cards now: 
        \begin{enumerate}[(a)]
            \item E
            \item K
            \item 4
            \item 7
        \end{enumerate}

        Which of these cards must in any case be turned over to check the rule? (In other words: which cards could violate the rule above?) 
        \begin{enumerate}[(a)]
            \item E
            \item K
            \item 4
            \item 7
        \end{enumerate}
 & \cite{wason_1966, bruckmaier_2021} \\ 
 \hline
        Wason task (facilitated) & Imagine that you are working for the post office. You are responsible for checking whether the right stamp is affixed to a letter. The following rule applies:
        
        If a letter is sent to the USA, at least one 90-cent stamp must be affixed to it.
        
        There are four letters in front of you, of which you can see either the front or the back.
        \begin{enumerate}[(a)]
            \item Letter 1: 90-cent stamp on the front
            \item Letter 2: Italy marked on the back
            \item Letter 3: 50-cents stamp on the front
            \item Letter 4: USA marked on the back
        \end{enumerate}

        Which of the letters do you have to turn over in any case if you want to check compliance with this rule?
        \begin{enumerate}[(a)]
            \item Letter 1
            \item Letter 2
            \item Letter 3
            \item Letter 4
        \end{enumerate}
 & \cite{bruckmaier_2021} \\
 \hline
        AIDS task & The probability that someone is infected with HIV is 0.01\%. The test recognizes HIV virus with 100\% probability if it is present. So, the test is positive. The probability of getting a positive test result when you don’t really have the virus is only 0.01\%. 
        
        The test result for your friend is positive. The probability that your friend is infected with the HIV virus is therefore: \_\_\_ \% 
        & \cite{bruckmaier_2021} adapted from \cite{eddy_1982} \\
        \hline
        AIDS task (facilitated) & This task involves an assessment of the results of the AIDS test. It is known that HIV can cause AIDS. Now imagine the following: A friend of yours gave blood at the hospital. It will then be checked to see if HIV is present in the blood. The test result is positive. How likely is it that your friend is actually infected with the HIV? 
        
        To answer this question, you will need the following information: Out of 10,000 people, 1 person is infected with HIV. If the person is infected with the HIV, the test detects HIV. So the test is positive. Only 1 of the 9,999 people who are not infected with HIV have a positive test.

        The test result for your friend is positive. How many people who have received a positive test result are actually infected with HIV? \_\_\_ from \_\_\_. & \cite{bruckmaier_2021} \\
        \hline
        Hospital problem & In hospital A about 100 children are born per month. In hospital B about 10 children are born per month. The probability of the birth of a boy or a girl is about 50 percent each. 
        
        Which of the following statements is right, which is wrong? 
        
        The probability that once in a month more than 60 percent of boys will be born is…
        \begin{enumerate}[(a)]
            \item … larger in hospital A
            \item … larger in hospital B
            \item … equally big in both hospitals
        \end{enumerate}
 & \cite{bruckmaier_2021}, adapted from \cite{kahneman_tversky_1972, tversky_kahneman_1974} \\
 \hline
        Monty Hall problem & A candidate on a quiz show can choose one of three doors. Behind one of the doors is the main prize, a car. Behind the other two doors, there are two goats. The rules of the game are now as follows: The quizmaster knows behind which of the doors the car and the goats are. After the candidate has chosen one of the doors, it remains locked for the time being. 
        
        The quizmaster then opens one of the other two doors. He always opens a door with a goat behind it. Imagine that the candidate chooses door 1. Instead of opening this door, the quizmaster opens another door, behind which there is a goat. He now offers the candidate the option of switching his choice to the last unopened door. 
        
        Should the candidate switch to the door or not? & \cite{friedman_1998, bruckmaier_2021}\\
        \hline
        
        Monty Hall problem (facilitated) & A candidate on a quiz show can choose one of three doors. Behind one of the doors is the main prize, a car. Behind the other two doors, there are two goats. The rules of the game are now as follows: The quizmaster knows behind which of the doors the car and the goats are. After the candidate has chosen one of the doors, it remains locked for the time being. The quizmaster then opens one of the other two doors. He always opens a door with a goat behind it. 
        
        Imagine that the candidate chooses door 1. Instead of opening this door, the quizmaster opens another door, behind which there is a goat. He now offers the candidate the option of switching his choice to the last unopened door. Should the candidate switch to the door or not?
        
        There are only three possible car-goat constellations:
        \begin{enumerate}[(a)]
            \item Door 1: goat (first choice); door 2: goat; door 3: car.
            \item Door 1: goat (first choice); door 2: car; door 3: goat.
            \item Door 1: car (first choice); door 2: goat; door 3: goat.
        \end{enumerate}
        Now think about in which constellation the quizmaster could open which door. Then decide whether the candidate should stay in the respective constellation with his first choice or switch to the last remaining door.
        
        In how many of these three constellations should the candidate switch from his first choice to the remaining unopened door? In \_\_\_ of 3 constellations. 
        
        What should the candidate therefore do? Stay or switch? & \cite{bruckmaier_2021} \\
        \hline
       
        Linda 
        
        problem & Linda is 31 years old, single, very intelligent, and speaks her mind openly. She studied philosophy. During her studies, she dealt extensively with questions of equality and social justice and participated in anti-nuclear demonstrations. 
        
        Now order the following statements about Linda according to how likely they are. 
        
        Which statement is more likely?
        \begin{enumerate}[(a)]
            \item Linda is a bank clerk.
            \item Linda is active in the feminist movement.
            \item Linda is a bank clerk and is active in the feminist movement.
        \end{enumerate}
 & \cite{tversky_kahneman_1983} \\
 \hline
        Birth 
        
        sequence problem (random) & All families with six children in a city were surveyed. In seventy-two families, the exact order of births of boys (B) and girls (G) was GBGBBG. What is your estimate of the number of families surveyed in which the exact order of births was BGBBBB? & \cite{kahneman_tversky_1972} \\
        \hline
        
        Birth 
        
        sequence problem (ordered)  & All families with six children in a city were surveyed. In seventy-two families, the exact order of births of boys (B) and girls (G) was GBGBBG. What is your estimate of the number of families surveyed in which the exact order of births was BBBGGG? & \cite{kahneman_tversky_1972} \\
        \hline
        High school problem & There are two programs in a high school. Boys are a majority (65\%) in program A, and a minority (45\%) in program B. There is an equal number of classes in each of the two programs. 
        
        You enter a class at random, and observe that 55\% of the students are boys. What is your best guess – does the class belong to program A or to program B? & \cite{kahneman_tversky_1972} \\
        \hline
        Marbles task & On each round of a game, 20 marbles are distributed at random among five children: Alan, Ben, Carl, Dan, and Ed. Consider the following distributions:
        
        Type I: Alan: 4; Ben: 4; Carl: 5; Dan: 4; Ed: 3.
        
        Type II: Alan: 4; Ben: 4; Carl: 4; Dan: 4; Ed: 4.
        
        In many rounds of the game, will there be more results of type I or of type II? & \cite{kahneman_tversky_1972} \\
        \hline
    \caption{Task Description}
    \label{task_desc}
\end{longtable}

\end{document}